\documentclass[letterpaper, 10 pt, conference]{ieeeconf}  

\usepackage{graphicx}
\usepackage{booktabs}
\usepackage{float}
\usepackage{amssymb,amsfonts}
\usepackage{listings}
\usepackage{url}
\usepackage{algorithm}
\usepackage{algorithmic}
\usepackage{amsmath,bm}
\usepackage[table,xcdraw]{xcolor}
\usepackage{scalerel}
\usepackage{tikz}
\usepackage{stfloats,cite}
\usepackage{multirow}
\usepackage{threeparttable}
\usepackage{nomencl}
\usepackage{comment}
\makenomenclature

\usepackage{hyperref}
\hypersetup{
    colorlinks=true,
    linkcolor=blue,
    filecolor=black,      
    urlcolor=black,
    citecolor=purple
}

\newcommand{\bbb}{{\rm{b}}}
\newcommand{\nnn}{{\rm{n}}}
\newcommand{\mmm}{{\rm{LIDAR}}}

\usepackage{fontawesome}

\makenomenclature

\IEEEoverridecommandlockouts                              

\title{\LARGE \bf
Incorporating GNSS Information with LIDAR-Inertial Odometry for Accurate Land-Vehicle Localization
}

\author{Jintao Cheng$^1$, Bohuan Xue$^1$, Shiyang Chen$^2$, Qiuchi Xiang$^1$ and Xiaoyu Tang$^1$\textsuperscript{\faEnvelopeO} 
\thanks{$^{1}$ South China Normal University
        }%
\thanks{$^{2}$ Charles III University of Madrid
        }%
\thanks{\emph{Corresponding author: Xiaoyu Tang: {\tt\small tangxy@scnu.edu.cn}}.}
}

\begin{document}

\maketitle
\thispagestyle{empty}
\pagestyle{empty}

\begin{abstract}

Currently, visual odometry and LIDAR odometry are performing well in pose estimation in some typical environments, but they still cannot recover the localization state at high speed or reduce accumulated drifts. In order to solve these problems, we propose a novel LIDAR-based localization framework, which achieves high accuracy and provides robust localization in 3D pointcloud maps with information of multi-sensors. The system integrates global information with LIDAR-based odometry to optimize the localization state. To improve robustness and enable fast resumption of localization, this paper uses offline pointcloud maps for prior knowledge and presents a novel registration method to speed up the convergence rate. The algorithm is tested on various maps of different data sets and has higher robustness and accuracy than other localization algorithms.

\end{abstract}


\section{Introduction}

Accurate localization is a crucial component of Autonomous driving\cite{MF,BM}. Besides integrated navigation-based solutions, the main approaches include LIDAR-based localization \cite{9636501,9197450,2014LOAM} and Vision-based localization \cite{7782863,7946260,8421746}. LIDAR-based schemes are mainly based on the point-cloud registration, which can be solved via iterative closest point (ICP) method \cite{Besl1992}. The ICP algorithm is effective, but may suffer from local minima in the face of sophisticated environments \cite{Yang2020}. Integrated Navigation-based methods can obtain high-precision global localization, while it often fails in complex urban environments. Furthermore, vision-based solutions are usually used in indoor environments because cameras are subject to interference from sunlight and rainy. Therefore, such methods may produce large cumulative errors. 

Among the different challenges involved, LIDAR-based localization exhibits high robustness and accuracy. In the case of pure LIDAR-based odometry,  the method in \cite{2014LOAM} scores among the highest in terms of translational and rotational errors on the Kitti Vision Benchmark \cite{6248074}. However, in the environment of long trajectory and geometric degradation, LIDAR odometry will generate accumulated errors and drift without a prior pointcloud map. In particular, other sensors need to be used to finish re-localization after the localization has been lost for a long time. On the other side, LIDAR registration  \cite{1992Iterative,1249285} (alignment with respect to the offline 3D pointcloud maps) can provide high-precision localization and solve the accumulated errors and drift to a certain extent, while common registration algorithms are difficult to realize real-time analysis \cite{8616217,8816641}.

\begin{figure*}[ht]
\centering
  \includegraphics[width=0.9\textwidth]{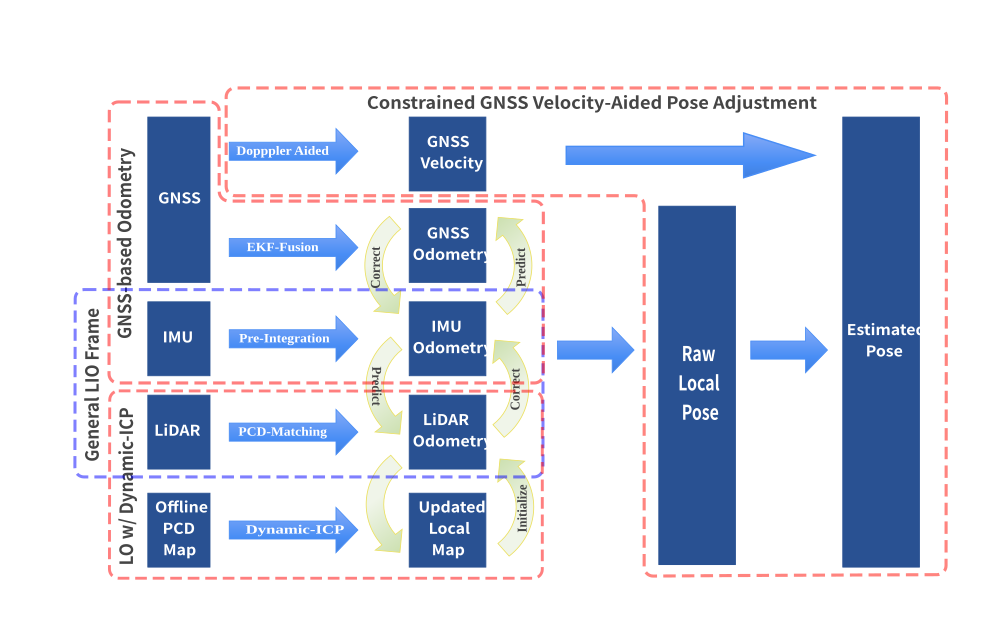}
  \caption{The diagram of our method scheme. The red dotted box shows our three main sub-frameworks, and the blue dotted box shows the general LIO frame.}
  \label{diagram}
\end{figure*}

Based on the above issues, we apply a sensor fusion framework to correct the accumulated error and accelerate the re-localization in the odometry. The framework realizes robust and high accurate localization between the online and a prior offline 3D pointcloud map. Compared with SLAM methods, our method uses the prior pointcloud map to make the whole localization system more robust. It also combines IMU with GNSS sensor to improve the global positioning accuracy and the relative positioning stability, so that it can compensate for the slow operation of the registration algorithm.

Our main contribution is proposing a robust sensor fusion localization framework, which can provide robust localization. In the complex urban environment, we perform pose optimization estimation by combining GNSS, IMU, and LIDAR. Compared with other localization frameworks, we use a prior map to complete pointcloud registration, which undoubtedly improves the stability of the system. Meanwhile, in order to improve the accuracy and efficiency of pointcloud registration, we propose an innovative registration algorithm Dynamic-ICP. Finally, we propose a mechanism using multi-sensor information to judge whether the registration fails. 

The main contributions of this paper are summarized as follows:
\begin{enumerate}
\item We present a robust localization system in a prior 3D feature map, which combines a GNSS-based global odometry and an optimized LIDAR-Inertial odometry.
\item We propose a novel Dynamic-ICP registration method, which is employed in the LIDAR odometry, to solve the re-localization problem and enhance localization accuracy.
\item We apply a constrained GNSS velocity-aided pose adjustment method to estimate the heading and help the system perform more accurate localization.
\end{enumerate}

\section{Related Work}
Both camera and LIDAR are fundamental sensors, which are crucial for autonomous localization problems. The relevant scientific research work can be divided into two sub-parts addressed in this paper. The characteristic of high frequency and low cost make Vision-based localization become one of the research directions. The works that are closely related to the method shown in this paper are LIDAR-based localization. In this section, we will briefly introduce state-of-the-art work in the field of Vision-based localization and LIDAR-based localization.

\subsection{Vision-based localization}
 Vision-based localization can be grouped under the three sub-methods including feature-based method, direct method, and semi-direct method. \cite{7946260} shows a feature-based method that uses various cameras and \cite{8421746} presents a robust and versatile monocular visual-inertial state estimator. Because the camera is easy to be affected by a wide range of illumination angles, the feature point method is not suitable for some repeated texture or weak texture environments. In \cite{DSO}, previous researchers propose a visual odometry based on a direct structure and motion formulation. Although it can solve the localization problem in a weak texture environment, its characteristics make it unable to complete re-localization. In \cite{7782863}, a semi-direct method, which uses intensity gradients to track and joint feature points optimization, has been proposed. Using different methods in visual odometry have various advantages, but it usually performs not well enough in complex environments.

\subsection{LIDAR-based localization}
In recent years, LIDAR-based approaches have developed rapidly. Apart from the classical point-to-point error terms, there have been many error metrics proposed for the ICP mapping. For instance, the point-to-line and point-to-plane metrics are now benefiting the robotics community from their highly accurate pointcloud mapping. However, these new metrics are more challenging since the induced problems are non-convex in terms of the rotation and translation, thus many efforts are paid to give efficient and accurate globally optimal solutions \cite{Wu2022tro}. Compared to classical algorithm\cite{2014LOAM,lego-loam,lio-sam} propose LIDAR Odometry based on graph optimization to achieve trajectory estimation and map-building. This method is generally used to build maps, but the computation is usually inefficient. \cite{DBLP:journals/corr/abs-2109-05483} propose a method of pre-tracker, which allows to obtain pre-computed odometry, to be used as aids when performing tracking. Pointcloud registration methods to estimate pose are presented in \cite{1992Iterative,1249285}. In \cite{9636501}, the authors introduce a fast direct localization in 3D pointcloud maps without features and point correspondences, while its method can not get the initial pose. \cite{9197450} shows a localization method joint LIDAR-Odometry \cite{2014LOAM} and scene recognition. Because segmentation is used in this algorithm, it needs to occupy computer graphics card resources. Despite the success of these methods, all the aforementioned methods do not complete the 
initial localization and re-localization on the pointcloud map, which means that their stability is questionable.

\subsection{Pointcloud Map based localization}
Based on the above method, some literatures use a prior map for localization. \cite{DSL,MVO-POINT} utilize features of the pointcloud maps in their visual system which can provide the accurate estimation of global 6-DoF camera poses with the absolute scale. Similarly, LIDAR-based localization makes use of the prior maps such as \cite{G-LOAM,9636501,9197450} to improve robustness. It can be seen that the constraints of pointcloud maps play a great role in improving the accuracy and robustness of localization.

\nomenclature{$\mathbb{R}^{n \times m}$}{the real space of the size $n \times m$}
\nomenclature{$\left \| \cdot \right \|$}{the Euclidean norm}
\nomenclature{$\bm{X}^\top$}{the transpose of an arbitrary real matrix $\bm{X}$}
\nomenclature{$[\bm{x} \times]$}{the skew symmetric matrix of a $3\times3$ vector $\bm{x}$}
\nomenclature{$\dot{\bm{x}}$}{the derivative of a vector $\bm{x}$ with respect to time $t$}
\nomenclature{$\bm{x} \left( m : n \right)$}{the vector with $m$-th to $n$-th elements of $\bm{x}$}
\nomenclature{$\phi$}{the Roll angle in the rotation mechanisms}
\nomenclature{$\theta$}{the Pitch angle in the rotation mechanisms}
\nomenclature{$\gamma$}{the Yaw angle in the rotation mechanisms}
\nomenclature{$\bm{R}_{\nnn}^{\bbb}$}{the rotation matrix from $\nnn$-frame to $\bbb$-frame}
\nomenclature{$\bm{t}$}{the translation matrix}
\nomenclature{$\bm{T}$}{the transformation matrix}
\nomenclature{$\bm{q}$}{the associated quaternion of rotation, subject to the unit constraint $\left \| \bm{q} \right \| = 1$}
\nomenclature{$\odot$}{the quaternion product}
\nomenclature{$\bm{p}$}{the pose matrix}
\nomenclature{$pts^{\bbb}$}{the LIDAR points in $\bbb$-frame}
\printnomenclature

\section{System Framework}

\subsection{System Review}

The flowchart of the proposed method is shown in Fig. \ref{diagram}. We take the LIDAR as the global coordinate system and convert the external parameters of GNSS and IMU to the LIDAR coordinate system according to the known parameters.

We introduce three odometry modules, including GNSS-based odometry, IMU-preintegration odometry, and LIDAR-based odometry into the localization framework. The GNSS-based Odometry provides global localization for initializing or restarting the entire system framework. The system uses Pre-integration odometry as motion prediction and joint LIDAR-based odometry while optimizing the poses. The LIDAR odometry employs our dynamic-ICP algorithm for initial localization or re-localization, but in other cases, the IMU prior information is utilized to accelerate the iterative optimization of ICP.

Meanwhile, this framework relies on a prior 3-D pointcloud map, which supports a variety of mapping algorithms. In the process of localization, the algorithm generates local maps according to the current pose set to improve the matching efficiency. Then, the robustness of frame localization in various algorithms will be presented in the next part.

\subsection{GNSS-based Odometry} 

 Although GNSS is greatly affected by the urban environment, it can supply reliable global information for the system. Compare to the optimization-based method\cite{lio-mapping}\cite{lio-sam}, filter-based method\cite{8967880}\cite{8461224} have the advantages of small amount of calculation and strong real-time performance. Since global odometry is only used for system initialization or restart, we use EKF\cite{2012Fundamentals} as the fusion method. It is a practical approach to use high-frequency IMU to complete the prediction process and low-frequency GNSS to correct.
 
\begin{align}
\bm{x} &= \left( \phi, \theta, \psi,
P_{\rm{e}},  P_{\rm{n}}, 
h_{\rm{MSL}},  
\bm{V}_{\rm{e}},  
\bm{V}_{\rm{n}}, 
\bm{V}_{\rm{u}}\right)^{\top} 
\end{align}
$\bm{x}$ is the output of state variance, where $\phi$, $\theta$ and $\psi$ represents three axis attitude angle. Let $P_{\rm{e}}$, $P_{\rm{n}}$, $h_{\rm{MSL}}$, $\bm{V}_{\rm{e}}$, $\bm{V}_{\rm{n}}$ and $\bm{V}_{\rm{u}}$ define as three axis pose and velocity in ENU coordinate system. 

\begin{align}
\bm{z} &= \left( 
\psi_{\rm {mag }},
P_{\rm{e, g}},
P_{\rm{n, g}},
h_{\rm{M S L, g}},
\bm{V}_{\rm{e, g}},
\bm{V}_{\rm{n, g}},
\bm{V}_{\rm{u, g}} \right)^{\top}
\end{align}
Where $\bm{z}$ is defined as the measurement variance, and $\psi_{\rm {mag}}$ is magnetometer measurement. GNSS provides there axis position $P_{\rm{e, g}}$, $P_{\rm{n, g}}$, $h_{\rm{M S L, g}}$ and three axis velocity $\bm{V}_{\rm{e, g}}$, $\bm{V}_{\rm{n, g}}$, $\bm{V}_{\rm{u, g}}$.

\subsection{IMU-preintegration Odometry}

We maintain an odometer based on IMU pre-integration\cite{2019supplementary} as the initial value of pointcloud registration. The raw measurements of the IMU are defined using Eqs. 1 and 2. 
\begin{align}
\hat{\bm{\omega}}_{k} &=\bm{\omega}_{k}+\bm{b}_{k}^{\bm{\omega}}+\bm{n}_{k}^{\bm{\omega}} \\
\hat{\bm{a}}_{k} &=\bm{R}_{k}^{\bm{-1} \bm{}}\left(\bm{a}_{k}-\bm{g}\right)+\bm{b}_{k}^{\bm{a}}+\bm{n}_{k}^{\bm{a}} 
\end{align}
where, letting the current timestamp as ${{k}}$, $\hat{\bm{\omega}}_{k}$ is the raw angular velocity with bias ${\bm{b}_{k}^{\bm{\omega}}}$ and noise ${\bm{n}_{k}^{\bm{\omega}}}$ in the time ${{k}}$; $\hat{\bm{a}}_{k}$ is the raw acceleration with bias ${\bm{b}_{k}^{\bm{a}}}$ and noise ${\bm{n}_{k}^{\bm{a}}}$; $\bm{{R}_{k}}^{\bm{-1}}$ is the rotation matrix from world coordinate system to body coordinate system; $\bm{g}$ is the constant gravity vector in world coordinate system.

Using the pre-integration method can obtain the relative body motion between two timestamps. The increments of displacement, velocity, and rotation in time ${i}$ and ${j}$ can be calculated as

\begin{align}
\Delta \bm{v}_{i j} &=\bm{R}_{i}^{\top}\left(\bm{v}_{j}-\bm{v}_{i}-\bm{g} \Delta k_{i j}\right) \\
\Delta \bm{p}_{i j} &=\bm{R}_{i}^{\top}\left(\bm{p}_{j}-\bm{p}_{i}-\bm{v}_{i} \Delta k_{i j}-\frac{1}{2} \bm{g} \Delta k_{i j}^{2}\right) \\
\Delta \bm{R}_{i j} &=\bm{R}_{i}^{\top} \bm{R}_{j}
\end{align}

Let ${v}_{k}$ defines as the velocity in time ${k}$, ${p}_{k}$ represents the displacement in time ${k}$, where $\bm{{R}_{k}}$ is the rotation in time ${k}$.

Due to the conversion $\bm{T}_{l2m}$ between the IMU coordinate system and the LIDAR coordinate system, we can define the pose of the IMU odometry in the LIDAR coordinate system $\bm{T}_{lidar}$ as

\begin{equation}
\begin{aligned}
\bm{T}_{lidar} &=\bm{T}_{l2m} \bm{T}_{imu}
\end{aligned}
\end{equation}

\subsection{LIDAR Odometry with Dynamic ICP}

LIDAR Odometry plays a significant role in the whole localization framework. Before starting the LIDAR Odometry process, we would like to load the offline map which is used the state of the art mapping algorithms\cite{lio-sam,2014LOAM,1249285}.

Algorithm \ref{alg:alg1} shows the Dynamic ICP Algorithm, which is a novel pointcloud registration algorithm to obtain fast and high accuracy localization in initial and re-localization. The odometry is calculated by the usage of a prior map and the current LIDAR frame, depending on the remarkable accuracy of ICP with a dynamic calculation and fast response. This method is able to apparently overcome the shortcomings of the normal ICP algorithm, the lack of costing such a long time, by setting the believable initial guess transformation between LIDAR frames and restricting the matching region as well. 

The process of the algorithm will be divided into two parts, region selecting and ICP matching. The first step is to select a region of the local map for further matching. Due to the lack of finding the absolute position in the map frame, it is such a hard challenge for LIDAR to find out where it is in the global map frame for the first time. A commonly used method to solve this problem is to match the frame with frames at plenty of way-points, saved while building the map to make sure the maximum likely pose in the map. However, it will cost a long time to search through all of the possible areas. On the contrary to this method, our algorithm takes usage of GNSS messages as an observation globally to offer the initial guess, and select a local map around the guess position with a smaller range that gives a higher success rate for each attempt. Moreover, the region size can be adaptive so as to make it effective in most scenes even if the GNSS data is not reliable enough. 
\begin{algorithm}
	\renewcommand{\algorithmicrequire}{\textbf{Input:}}
	\renewcommand{\algorithmicensure}{\textbf{Output:}}
	\caption{Dynamic ICP Algorithm}
	\label{alg:alg1}
	\begin{algorithmic}[1]
		\STATE Initialization: $\bm{T}_{k,0} = \left[ \bm{R}_{k,0}, \bm{t}_{k,0}; 0, 1 \right], i \leftarrow 0$ , \\ $r_{k} \leftarrow \rm{SCANRANGE}  $ (via GNSS)
		\STATE  ${\bm{p}_{k,0}} = \bm{T}_{k,0}\cdot \bm{p}_{k-1} $ \label{eqn_1}
		\IF{ $||\bm{pts}^{gmap}_n, \bm{p}_{k,0}|| < r_{k} $}
    		\STATE $\bm{pts}^{m}_k \leftarrow \bm{pts}^{gmap}_n$ \label{eqn_2}
		\ENDIF
		\STATE $E(\bm{T}_k) = \mathop{\arg\min}\limits_{\bm{T}_k \in SE3}\frac{1}{|pts^{l}|}\sum\limits_{n=1}^{pts^{l}} \bm{pts}^{m}_n-\left(\bm{t}_k + \bm{R}_k\cdot \bm{pts}^{l}_n \right)$ \label{eqn_3}
		\REPEAT
		\STATE $i \leftarrow i + 1$
		\STATE Update $ \bm{T}_{k,i} $ based on Equation~(\ref{eqn_3})
		\STATE Update $ \bm{p}_{k,i} $ based on Equation~(\ref{eqn_2})
		
		\UNTIL $E(\bm{T}_k) < \rm{Error \ Tolerance}$
		\STATE $\bm{T}_k = \bm{T}_{k,i}$
		\STATE $\bm{p}_k = \bm{p}_{k,i}$
		\ENSURE transformation matrix $\bm{T}_k$ \& current pose $\bm{p}_k$
	\end{algorithmic}  
\end{algorithm}

\begin{figure}[htbp]
  \centering
  \includegraphics[width=0.48\textwidth,
    trim={90 210 90 210},
    clip
  ]{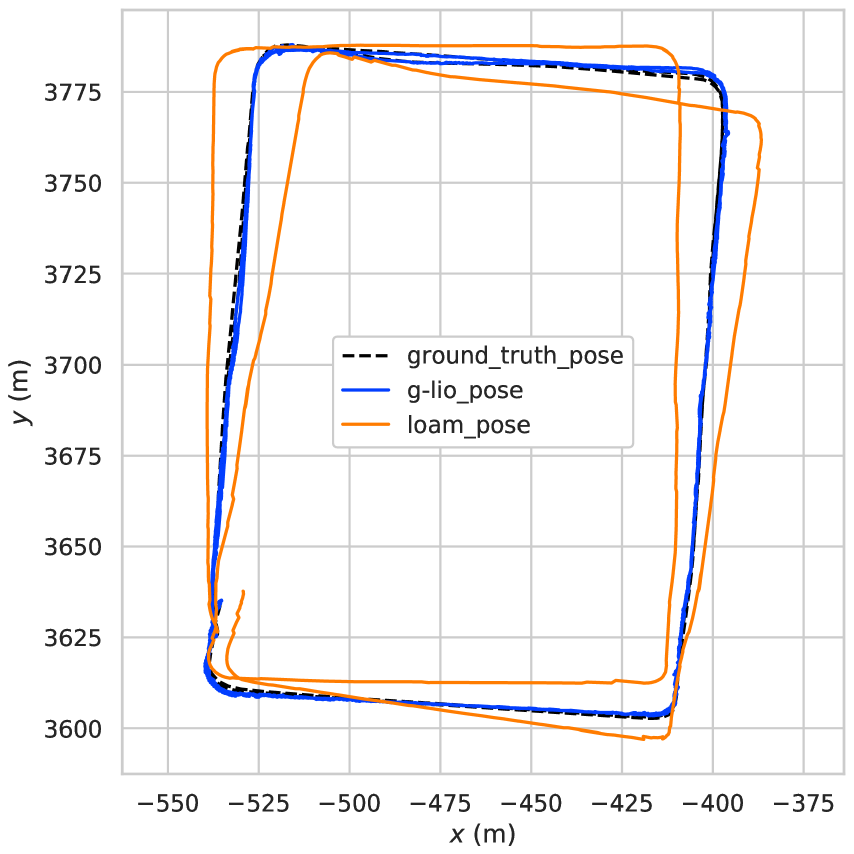}
  \caption{Comparision of trajectory results of our method, LOAM and ground truth in HK02 \cite{hsu2021urbannav} dataset.}
  \label{fig:multi-map}
\end{figure}

\subsection{Constrained GNSS Velocity-Aided Pose Adjustment}
We use an indirect Kalman filter (IKF) to estimate the heading, pitch angles along with non-gravitational acceleration with the GNSS velocity. It runs with a relatively low sampling rate (typical time period: 1 s). It substantially plays an important role in correcting systematic errors of MARG sensors. More importantly, the GNSS velocity vector contains information of pitch and yaw, with roll to be an undetermined stated in the Kalman filter. In this way, a constrained IKF is proposed to constrain the roll angle during the propagation of Kalman filtering. We use the error quaternion representation to propagate the IKF. The error quaternion is defined as ${{{\bm{\hat q}}}_{e,k}} = \left( 1, \bm{y}_k^\top \right)^\top$ in the time instant $k$, with error term $\bm{y}_k$ being the state vector of the IKF. The quaternion is propagated using.
\begin{eqnarray}
\label{equ41}
&&{{{\bm{\hat q}}}_{k + 1}} = \frac{{{{{\bm{\hat q}}}_{k}} \odot {{{\bm{\hat q}}}_{e,k}}}}{{\left\| {{{{\bm{\hat q}}}_{k}} \odot {{{\bm{\hat q}}}_{e,k}}} \right\|}}
\end{eqnarray}
With (\ref{equ41}), ${\bm{R}}_{{\nnn}}^{{\bbb}}$ at time $k$ is renewed with ${{\bm{\hat q}}_{k}}$ and the initial value of ${\bm{\hat q}}_{\xi,  k} $ is reset as zero. Thus for the IKF, its predicted state is computed by previous LIDAR odometry, with covariance of ${{\bm{P}}_{k}}$. The observation model is written as
\begin{eqnarray}
\label{equ44}
{{\bm{\tilde w}}_{v, k}}={{\bm{H}}_{v, k}}{{\bm{y}}_k}{\rm{ + }}{{\bm{\varepsilon }}_{{s}}}
\end{eqnarray}
where \[{{\bm{\tilde w}}_{v, k}} = {{\bm{w}}_{v, k}} - {\bm{R}}_{{\nnn}}^{{\bbb}}\left( {{{{\bm{\hat y}}}_{{{\mmm}},k}}} \right){{\bm{v}}^{\nnn}}\]
\[{{\bm{H}}_{v, k}} = 2 \left[ \left( {\bm{R}}_{{\nnn}}^{{\bbb}}\left( {{{{\bm{\hat q}}}_{{{\mmm}},k}}} \right){{\bm{v}}^{\nnn}} \right) \times \right ]. \]
However, as is well understood, velocity generated from a single antenna GNSS does not provide any roll information itself. In such point of view, we conduct a roll-constraint based IKF. Defining the IKF estimate of quaternion error as ${{\bm{\hat q}}_{e,v, k}}$, then the final rotation matrix is obtained through
\begin{eqnarray}
\label{equ45}
{\bm{R}}_{{\nnn}}^{{\bbb}}\left( {{{{\bm{\hat q}}}_{v, k}}} \right)={\bm{R}}_{{\nnn}}^{{\bbb}}\left( {{{{\bm{\hat q}}}_{e,v, k}}} \right){\bm{R}}_{\nnn}^{\bbb}\left( {{{{\bm{\hat q}}}_{k}}} \right)
\end{eqnarray}
where \[{\bm{R}}_{{\nnn}}^{{\bbb}}\left( {{{{\bm{\hat q}}}_{e,v, k}}} \right)=\left[ {\begin{array}{*{20}{c}}
	{\rm{1}}&{2\bm{ \hat y}_k\left( 3 \right)}&{ - 2\bm{ \hat y}_k\left( 2 \right)}\\
	{ - 2\bm{ \hat y}_k\left( 3 \right)}&1&{2\bm{ \hat y}_k\left( 1 \right)}\\
	{2\bm{ \hat y}_k\left( 2 \right)}&{ - 2\bm{ \hat y}_k\left( 1 \right)}&1
	\end{array}} \right]\] and ${{\bm{\hat y}}_{k}\left( i \right)}$ for $i = \left\{ {1,2,3} \right\}$ denotes the $i$-th element of IKF estimate ${{\bm{\hat y}}_{v, k}}$.\\
\indent Considering roll constraint of attitude quaternion, we have
\begin{eqnarray}
\label{equ46}
f\left( {{{{\bm{\hat q}}}_{e,v, k}}} \right) = \frac{{{{\left[ {{\bm{R}}_{{\nnn}}^{{\bbb}}\left( \bm{q}_k \right)} \right]}_{\left( {2,{\rm{3}}} \right)}}}}{{{{\left[ {{\bm{R}}_{{\nnn}}^{{\bbb}}\left( \bm{q}_k \right)} \right]}_{\left( {3,{\rm{3}}} \right)}}}} - \frac{{{{\left[ {{\bm{R}}_{{\nnn}}^{{\bbb}}\left( {{{{\bm{\hat q}}}_{\mmm,k}}} \right)} \right]}_{\left( {2,{\rm{3}}} \right)}}}}{{{{\left[ {{\bm{R}}_{{\nnn}}^{{\bbb}}\left( {{{{\bm{\hat q}}}_{\mmm,k}}} \right)} \right]}_{\left( {3,{\rm{3}}} \right)}}}}{\rm{ = 0}}
\end{eqnarray}
Referring to (\ref{equ46}), this constraint indicates that the IKF does not exert any roll correction on the previous solution. Inserting (\ref{equ45}) into (\ref{equ46}) yields an equality
\begin{eqnarray}
\label{equ47}
{{\bm{\Theta }}_k^\top}{{\bm{\hat y}}_{k}} = 0
\end{eqnarray}
where ${\bm{\Theta }}_k$ is given by (\ref{large_eq}).
\begin{figure*}[hb]

\end{figure*}
The following optimization problem is constructed with
\begin{equation}\label{equ48}
\begin{gathered}
\mathop {\arg \min }\limits_{{\bm{\hat y}}_k}  \frac{1}{2}{\bm{V}}_k^\top{\bm{\Sigma}}_{{v}}^{ - 1}{{\bm{V}}_k} ,\ \ \ \ {\rm{s.t.}} \ \ \ {{\bm{\Theta }}_k}^\top{{{\bm{\hat y}}}_{k}} = 0
\end{gathered}
\end{equation}
where ${{\bm{V}}_k}$ is the residual for observation model,
\begin{eqnarray}
\label{equ49}
&&{{\bm{V}}_k} = {{\bm{H}}_{v, k}}{{{\bm{\hat y}}}_{k}} - {{{\bm{\tilde w}}}_{v, k}}
\end{eqnarray}
Introducing the Lagrange multiplier $\lambda$ gives the Lagrangian of
\begin{equation}
{\mathcal{L}} = \frac{1}{2}{\bm{V}}_k^\top{\bm{\Sigma}}_{{v}}^{ - 1}{{\bm{V}}_k} + {\lambda}{{\bm{\Theta }}_k}^\top \bm{\hat y}_k
\end{equation}
Evidently, ${\mathcal{L}}$ is a convex and continuous function of $\bm{\hat y}_k$ and $\lambda$. Taking the derivative of ${\mathcal{L}}$ with respect to ${\bm{\hat y}}_k$ and $\lambda$, we obtain the gradient
\begin{equation} \label{equ50}
\begin{gathered}
\nabla \mathcal{L}_{\bm{y}} =   {{\bm{H}}_{{{v,}}k}^\top} {\bm{\Sigma}}_{{v}}^{ - 1} \left( {{{\bm{H}}_{{{v,}}k}}{{{\bm{\hat y}}}_{k}} - {{{\bm{\tilde w}}}_{{{v,}}k}}} \right) + {\lambda}{{\bm{\Theta }}_k} \hfill \\
\nabla \mathcal{L}_{\lambda} =  {{\bm{\Theta }}_k^\top}{{\bm{\hat y}}_{k}} \hfill
\end{gathered}
\end{equation}
Zeroing the gradient gives solution to $\bm{\hat y}_k$, which is in the form of a linear system that is easy to be solved.

\subsection{LIDAR-Inertial-GNSS Integrated Navigation}
To solve the drawback of low frequency LIDAR Odometry, jointing IMU Odometry is an efficient method. We use the quadratic curve approximate interpolation method to fuse the output of IMU and LIDAR. Let the result output is $Q^{l}$. 

In order to enhance the system robustness, we utilize GNSS information to optimize. The optimization formula is as follows, where $\alpha$, $\beta$ and $A$ are super parameters. Let $\bm{V}_{g}$ defines as GNSS velocity, where ${\Delta \bm{Q}}_{l}$ is the relative pose between timestamp $t$ and $t-1$,  $\bm{Q}_t$ and $\bm{Q}_{t-1}$ is the IMU and lidar fused output in timestamp $t$ and $t-1$. ${\bm{\Xi}}$ represents the matrix of GNSS covariance. $\bm{Q}_t^{g}$ is defined as GNSS-based Odometry in the timestamp $t$.

\begin{equation}
\hat{\bm{Q}}= \begin{cases}(1-\beta) \int_{\Delta t} \bm{V}_{g}+\beta {\Delta \bm{Q}^{}}^{l}+\bm{Q}_{t-1}^{l} & \operatorname{tr}({\bm{\Xi}})>A \\ (1-\alpha) \bm{Q}_{t}^{g}+\alpha \bm{Q}_{t}^{\prime} &\operatorname{tr} ({\bm{\Xi}}) \leq A\end{cases}
\end{equation}

The trace of the covariance matrix of GNSS reflects the current position degradation degree to a certain extent. When the trace is greater than {A}, the GNSS position information has a certain degradation, and the effectiveness of the current GNSS based data is reduced, but the velocity information integral of GNSS can be used for fusion. When the trace is less than {A}, the fusion of GNSS odometry can well correct the cumulative drift of odometry.

\section{Experimental Results}

We carry out experiments on a system equipped with an Intel i7-9700 processor, with 32Gb of RAM in all experiments. Prior pointcloud maps are built in all sequences with various mapping algorithms to test localization accuracy. 

Here are our test results on public dataset HK01, HK02 \cite{hsu2021urbannav}, and KITTI18, KITTI28 \cite{6248074}. We tested our algorithm on these dataset with the comparison of other widely used SLAM methods, including \cite{2014LOAM}\cite{Besl1992}\cite{1249285}. The mainly used indicator for evaluating the efficiency in this section is Root Mean Square Error(RMSE). The results show that our method performs better than the base methods, and the data of experiments are shown detailedly in TABLE I.

\begin{table}[]
\caption{RMSE experiment results table on different dataset}
\begin{center}
\begin{threeparttable}
\begin{tabular}{cccccc}
\hline
\hline
Sequence                             & MAP\_METHOD & LOAM\tnote{1}                  & ICP  & NDT  & OURS          \\ \hline
\multirow{2}{*}{\textit{HK01}}    & NDT-MAPPING & \multirow{2}{*}{4.57} & 2.00    & 3.13 & \textbf{1.98}  \\
                                     & LEGO\_LOAM  &                       & 4.34 & 5.50  & \textbf{1.96}  \\ \hline
\multirow{2}{*}{\textit{HK02}}     & NDT-MAPPING & \multirow{2}{*}{31.69}     & 0.88 & 0.92 & \textbf{0.77}  \\
                                        & LEGO\_LOAM  &                       & 1.17 & 1.13 & \textbf{1.08} \\ \hline
\multirow{2}{*}{\textit{KITTI18}}& NDT-MAPPING & \multirow{2}{*}{18.00}     &   X\tnote{2}   & 3.62 & \textbf{0.99}  \\
                                         & LEGO\_LOAM  &                       &   X   & 3.92 & \textbf{1.22}  \\ \hline
\multirow{2}{*}{\textit{KITTI28}}  & NDT-MAPPING & \multirow{2}{*}{32.00}     & X    & 9.66 & \textbf{1.98}  \\
                                        & LEGO\_LOAM  &                       & X    & 8.00    & \textbf{2.24}  \\ \hline 
\hline
\end{tabular}
\begin{tablenotes}
        \footnotesize
        \item[1] The LOAM method used odometers directly and was independent of how the map is constructed.
        \item[2] The ICP method does not run completely on the KITTI dataset, so its RMSE is not calculated in this paper. 
      \end{tablenotes}
\end{threeparttable}
\end{center}
\end{table}

\section{Conclusions}
At high speeds, commonly used LIDAR and visual odometry systems produce accumulated drift that is difficult to eliminate. Additionally, there has been a lack of fast re-localization methods when the predicted pose is inaccurate. Our proposed framework introduces three odometry modules, including GNSS-based odometry, which provides reference values for LIDAR-inertial odometry at a relatively low sampling frequency by introducing GNSS velocity and pose information, as well as IMU pre-integrated odometry and LIDAR-based odometry. Moreover, in order to solve the re-localization problem, we propose the Dynamic-ICP method, which selects the most likely suitable size area in the local map through the area selection algorithm to improve the convergence speed of point cloud registration and re-localization accuracy. In other cases, The system uses pre-integrated odometry as motion prediction to joint lidar-based odometry while optimizing poses, and IMU priors are used to speed up the iterative optimization of pointcloud matching. Since an offline 3-D pointcloud map is needed by our framework, it is designed to adapt maps generated by the most commonly used mapping algorithms. Through experimental comparisons, it can be seen that the algorithm remains stable in various complex environments and both single-Lidar and multi-Lidar platforms. In future work, we plan to add more sensor information, such as GNSS raw data, to update our positioning framework.



\bibliographystyle{IEEEtran}
\bibliography{root}

\end{document}